\pdfoutput=1

\documentclass[11pt]{article}

\usepackage[final]{acl}

\usepackage{times}
\usepackage{latexsym}

\usepackage[T1]{fontenc}

\usepackage[utf8]{inputenc}

\usepackage{microtype}

\usepackage{inconsolata}

\usepackage{graphicx}
\usepackage{amsmath, amssymb}
\usepackage{booktabs}
\usepackage{multirow}
\usepackage{tabularx}
\usepackage{soul}
\usepackage{enumitem}
\usepackage{placeins}
\usepackage{algorithm}
\usepackage[noend]{algpseudocode}
\usepackage{pgfplots}
\usepgfplotslibrary{groupplots}
\pgfplotsset{compat=1.18}
\usepackage{tikz}
\usepackage{arydshln}

\newcolumntype{L}{>{\raggedright\arraybackslash}X}
\newcolumntype{R}{>{\raggedleft\arraybackslash}X}
\newcolumntype{C}{>{\centering\arraybackslash}X}

\title{Hint-Augmented Re-ranking: Efficient Product Search using LLM-Based Query Decomposition}

\author{
  Yilun Zhu ~~~ ~~~ Nikhita Vedula ~~~  ~~~ Shervin Malmasi \\
  Amazon.com, Inc. \\
  \texttt{\{yilunzhu,veduln,malmasi\}@amazon.com}
}

\begin{document}
\maketitle
\begin{abstract}
Search queries with superlatives (e.g., \emph{best}, \emph{most popular}) require comparing candidates across multiple dimensions, demanding linguistic understanding and domain knowledge. We show that LLMs can uncover latent intent behind these expressions in e-commerce queries through a framework that extracts structured interpretations or \textit{hints}. Our approach decomposes queries into attribute-value hints generated concurrently with retrieval, enabling efficient integration into the ranking pipeline. Our method improves search performanc eby 10.9 points in MAP and ranking by 5.9 points in MRR over baselines. Since direct LLM-based reranking faces prohibitive latency, we develop an efficient approach transferring superlative interpretations to lightweight models. Our findings provide insights into how superlative semantics can be represented and transferred between models, advancing linguistic interpretation in retrieval systems while addressing practical deployment constraints.
\end{abstract}

\section{Introduction}
Effective document ranking requires sophisticated reasoning, particularly when relevance judgments depend on domain knowledge and logical inference \citep{niu2024judgerankleveraginglargelanguage, su2025brightrealisticchallengingbenchmark, ji2025reasoningrankteachingstudentmodels}. Although LLMs excel at reasoning-based tasks \citep{zhuang2024setwise, qin-etal-2024-large, chen2025attentionlargelanguagemodels}, their computational cost and inference latency remain major barriers to deployment in high-traffic environments \citep{chen2024frugalgpt}.
E-commerce search represents a particularly challenging scenario, where user queries frequently include superlative expressions (e.g., \textit{best}) that implicitly delegate the burden of domain-specific evaluation to the search engine. Such queries often arise when users lack detailed domain expertise; for example, a user unfamiliar with balloon types might search for ``\textit{best balloons for birthday parties}'' rather than specifying attributes explicitly, such as color variation, durability, material quality, or age appropriateness.

Interpreting these superlative expressions poses a fundamental linguistic and semantic challenge as it requires decomposing abstract comparative terms into concrete, context-specific attributes, a task demanding both semantic reasoning and domain-specific knowledge. Previous research highlights that interpreting superlatives involves commonsense reasoning \citep{bos-nissim-2006-empirical, scheible-2007-towards}, and remains difficult even for advanced LLMs \citep{pyatkin-etal-2025-superlatives}. Although prior work has explored superlatives in dialogue state tracking and product review analysis \citep{zhang-etal-2015-semantic,bakhshandeh-etal-2016-learning}, interpreting superlative queries for product search remains under-explored, despite its practical importance.

Recent studies have leveraged LLM-generated query expansions (QE) primarily to enhance retrieval effectiveness \citep{wang-etal-2023-query2doc, jagerman2023queryexpansion, dhole2024genqrensemble}, without explicitly addressing the decomposition of superlatives into concrete attributes. Moreover, existing QE methods have largely overlooked evaluating their impact on the ranking stage. Meanwhile, superlative queries are becoming increasingly common in evaluation frameworks \citep{dhole-etal-2025-generative}, underscoring this research gap. We fill this gap by explicitly decomposing superlatives into structured, domain-specific attributes (\textit{hints}), showing benefits in both retrieval and reranking stages. We focus on efficiency, enabling parallel execution of retrieval and hint generation to meet latency constraints.

Our contributions in this paper are as follows: 

    (i) We curate a dataset of 21k queries and 470k products for evaluating superlative query understanding in product search.
    
    (ii) We propose a method that efficiently leverages LLM reasoning abilities by extracting domain-specific knowledge and criteria as \textit{hints} from superlatives to guide lightweight re-ranking systems.
    
    (iii) Our approach enhances retrieval quality and enables small language models to surpass the performance of large models with faster inference.

Overall, our findings demonstrate how explicit superlative interpretation bridges the gap between user preferences expressed in natural language and practical search systems, efficiently integrating advanced reasoning into real-world applications\footnote{Code and data is available at: \url{https://github.com/yilunzhu/superhints/}}.

\begin{figure*}[t!h]
    \centering
    \vspace{-0.3cm}
    \includegraphics[height=5.5cm]{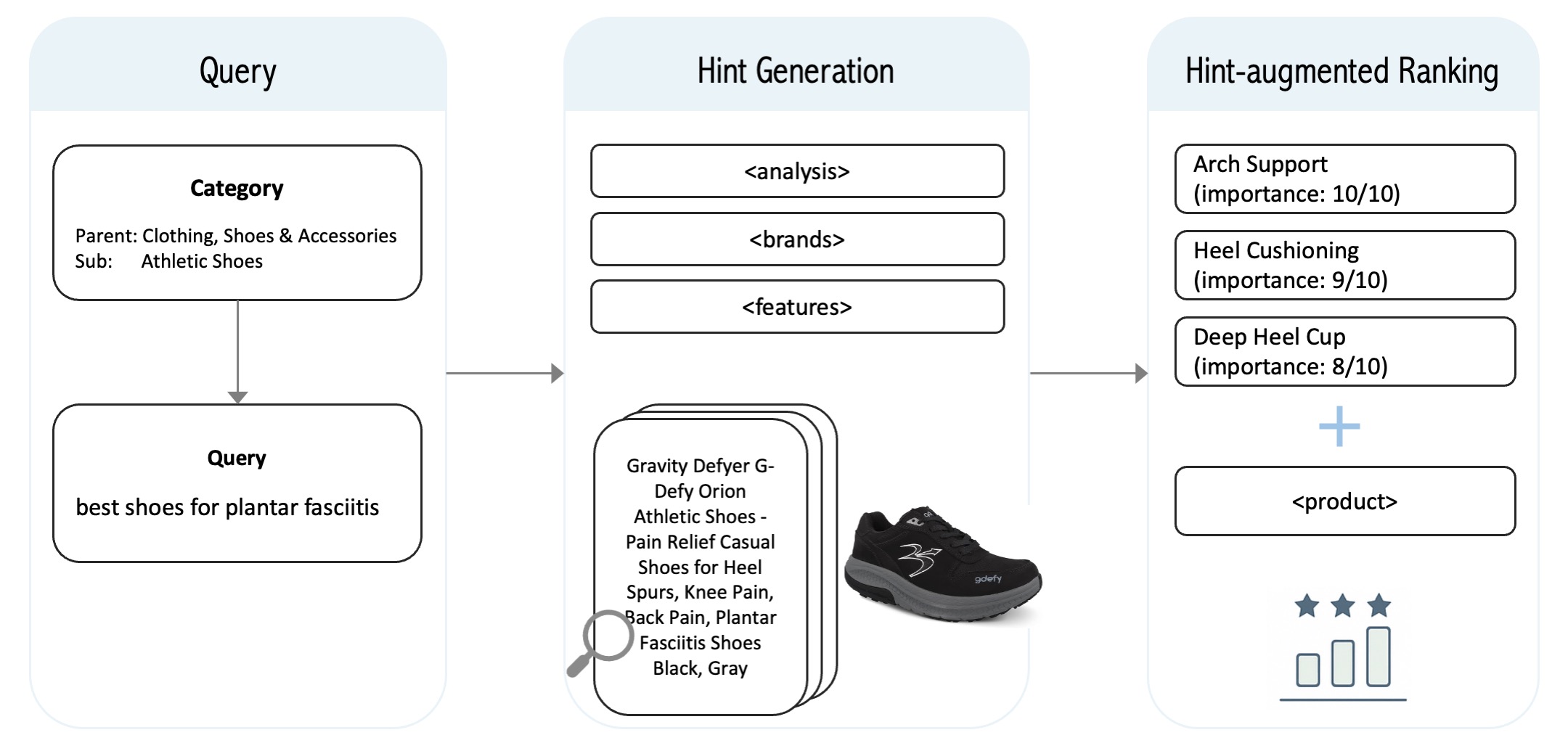}
    \vspace{-0.1cm}
    \caption{The framework of our proposed hint-augmented ranking approach.}
    \label{fig:framework}
\end{figure*}

\section{Related Work}
\paragraph{Superlatives}
Research on superlative interpretation in NLP has established that understanding these expressions requires commonsense knowledge about contextual comparison standards \citep{bos-nissim-2006-empirical, scheible-2007-towards}. While superlative interpretation has been applied to dialogue state tracking and product review mining \citep{bakhshandeh-etal-2016-learning}, and learned through structured knowledge bases \citep{zhang-etal-2015-semantic}, recent work shows they remain challenging even for advanced LLMs like GPT-4 \citep{pyatkin-etal-2025-superlatives}. Although evaluation frameworks have begun incorporating superlatives in test queries \citep{dhole-etal-2025-generative}, there remains a significant gap in addressing their interpretation specifically in search contexts where users frequently employ terms like ``best'' when lacking domain expertise, making accurate interpretation critical for relevant results.

\paragraph{LLM Ranking}
LLMs have recently emerged as powerful zero-shot rankers, significantly impacting information retrieval systems. Previous work demonstrated LLMs' capability to perform list-wise or pairwise relevance judgments without task-specific fine-tuning, achieving state-of-the-art performance on established information retrieval benchmarks~\cite{sun-etal-2023-chatgpt, pradeep2023rankvicunazeroshotlistwisedocument, qin-etal-2024-large,senel2024generative}. 

\paragraph{Datasets for Information Retrieval}
Existing information retrieval (IR) datasets, such as MS MARCO \citep{DBLP:journals/corr/NguyenRSGTMD16} and TREC \citep{li-roth-2002-learning}, are widely used for benchmarking retrieval methods. However, these datasets contain very few superlative queries (e.g., fewer than 500 in TREC), severely limiting their suitability for specifically studying superlative interpretation. While evaluating on traditional IR benchmarks is valuable, our work explicitly addresses superlative query understanding, necessitating datasets with sufficient coverage of these linguistic phenomena. To overcome this limitation, we adopt the publicly available Amazon Shopping Queries Dataset from the KDD 2022 Cup, which provides extensive coverage of superlative expressions in search contexts.

\section{Dataset Construction}

\subsection{Superlative Queries}
Despite the frequency of superlative queries in product search, existing datasets contain relatively few examples. Although \citet{dhole-etal-2025-generative} developed a superlative query dataset, we could not use it due to two limitations: its small scale (2.2K queries) provides insufficient training data, and its reliance on the original ESCI labels \citep{reddy2022shopping} excludes many potentially relevant products from the complete catalog, preventing accurate ranking evaluation. We therefore construct a specialized superlative query dataset using products from the Amazon Shopping Queries Dataset \citep{reddy2022shopping}. First, we extract two-level (parent and sub) category information for each product, excluding subcategories with fewer than 300 products to ensure sufficient diversity. For each qualifying subcategory, Claude 3.7 Sonnet \citep{claude3.7sonnet} generated 50 superlative queries, filtering those that retrieved no results with BM25. This process yields 26,993 candidate queries. A list of sample queries is in Appendix \ref{appendix:example_query}.

\begin{table}[t!]
\centering\small
\begin{tabular}{l|r|r|c|c|c}
\toprule
\multicolumn{1}{c|}{\multirow{2}{*}{\textbf{Split}}} & \multirow{2}{*}{\textbf{\# Pairs}}  & \multicolumn{1}{l|}{\multirow{2}{*}{\textbf{\# Queries}}} & \multicolumn{3}{c}{\textbf{Label Distribution (\%)}} \\ 
\cline{4-6} 
\multicolumn{1}{c|}{} & & \multicolumn{1}{l|}{} & \multicolumn{1}{l|}{\rule{0pt}{3ex}R\&B} & \multicolumn{1}{l|}{R-NB} & IRR   \\
\midrule
Train & 695k & 13,914 & 14.7\% & 56.3\% & 29.0\% \\
Dev   & 107k & 2,140 & 14.7\% & 56.0\% & 29.4\% \\
Test  & 267k & 5,353 & 14.7\% & 56.2\% & 29.1\% \\
\midrule
Total & 1,070k & 21,407 & 14.7\% & 56.3\% & 29.1\% \\
\bottomrule
\end{tabular}
\caption{Dataset statistics. R\&B: \textsc{relevant and best}, R-NB: \textsc{relevant but not best}, IRR: \textsc{irrelevant}.}
\label{tab:dataset}
\end{table}

\subsection{Relevance Labels}
To create relevance judgments, we first apply a dense retriever \texttt{gte-Qwen2-7B-instruct} \citep{li2023towards} to retrieve the top 50 products for each query. We then filter out queries with either no relevant products or too many relevant products ($\geq 15$), reducing our dataset to 21,407 queries.
Each query-product pair, we assign one of three labels (detailed definitions in Appendix \ref{appendix:prompts}):
\textsc{relevant and best} (product meets requirements and excels in dimensions implied by the superlative), \textsc{relevant but not best} (product meets requirements but doesn't excel in superlative dimensions), and \textsc{irrelevant} (product doesn't meet query requirements).

The final dataset contains 1,070,350 query-item pairs spanning 56 unique parent categories. We performed quality evaluation of the LLM-generated queries (Appendix \ref{appendix:data_quality}) and split the dataset using a stratified process based on label counts and parent category distribution, as shown in Table \ref{tab:dataset}.

\subsection{Decomposing Superlatives into Hints}
\label{sec:hint_gen}

We investigate the transfer of world knowledge from large-scale LLMs to smaller language models and statistical retrieval systems. To accomplish this, we employ Claude 3.5 Sonnet v2 \citep{claude3.5sonnet} to interpret superlative terms within the context of specific queries and product categories. 
For each query, the model generates four structured information blocks using Chain-of-Thought (CoT) reasoning \citep{wei2022chain}: (1) reasoning -- a detailed explanation of what the superlative term means in the given context; (2) brands -- a list of suggested brands known for excellence in the relevant category; (3) features -- key product attributes that correspond to the superlative term; and (4) generated queries -- alternative formulations that capture the query intent without superlative terms.

These generated \textit{hints} serve multiple purposes: the reasoning provides interpretable explanations of superlative terms, suggested brands inform retrieval for unfamiliar products, extracted features guide attribute-based filtering, and the alternative query formulations enhance retrieval coverage. Appendix \ref{appendix:prompts} shows an example. We generate these hints offline instead of using reasoning models at inference time, significantly reducing latency while preserving knowledge transfer benefits.\footnote{Reasoning models are excluded during inference due to their excessive latency, making them impractical in real-time.}
Our hint-augmented framework is presented in Figure \ref{fig:framework}.

\section{Experiments \& Evaluation}
\subsection{Retriever Models}
\paragraph{Sparse Retrievers}
We implement two sparse retrievers: a vanilla BM25 and Query Enhanced BM25 (QE-BM25). The vanilla BM25 implementation directly utilizes the original query including superlative terms without modification, serving as our baseline sparse retrieval method.

Our proposed QE-BM25 enhances the vanilla system by incorporating LLM-generated queries along with suggested brands from the hint generation step based on the Quasi Round Robin BM25 from \citet{li-2024-calrec}. For each original superlative query, we use the multiple queries generated by the LLM in Section~\ref{sec:hint_gen} to perform separate BM25 retrievals. We calculate the BM25 score for the top 50,000 items for each generated query. This threshold is determined via empirical analysis to ensure comprehensive coverage while maintaining computational efficiency (details in Appendix \ref{appendix_subsec:coverage}). Additionally, we also include doc2query \citep{doc2query_gospodinov_2023}, a query expansion baseline expanding query via LLMs and inputs to sparse retrieval.

By leveraging multiple LLM-generated queries, we aim to use the model's world knowledge to better interpret superlative terms. Unlike \citeauthor{li-2024-calrec}'s QRR-BM25, which assigns weight scores based on language modeling loss, we treat each generated query equally since commercial LLM logits are inaccessible. For each of the 10 generated queries, we compute BM25 scores for top products, store results in a matrix, and average scores to produce the final ranked list (algorithm in Appendix \ref{appendix_subsec:algorithm}).

\begin{table*}[t!]
\centering\small
\begin{tabular*}{\textwidth}{@{\extracolsep{\fill}}l|l|cccccc}
\toprule
\textbf{Retriever} & \textbf{Re-ranker} & \textbf{P@1} & \textbf{P@3} & \textbf{P@5} & \textbf{P@10} & \textbf{MAP} & \textbf{MRR} \\
\midrule
BM25 & \multirow{3}{*}{None} & 14.91 & 14.10 & 13.34 & 12.41 & 24.79 & 26.95 \\
doc2query & & 15.73 & 15.84 & 15.17 & 14.01 & 27.84 & 30.27 \\
QE-BM25 & & \underline{25.93} & \underline{24.60} & \underline{23.24} & \underline{21.01} & \underline{35.74} & 38.74 \\
\texttt{gte-qwen2-7b-instr} & & 20.14 & 22.27 & 20.48 & 19.06 & 35.01 & \underline{40.15} \\
\midrule
\multirow{6}{*}{\texttt{gte-qwen2-7b-instr}} & \texttt{qwen2.5-72B-instr} + Listwise & 35.31 & 35.63 & 35.05 & 31.08 & 25.81 & 55.02 \\
& \texttt{deepseek-r1} + Listwise & 44.35 & 42.59 & 39.95 & 34.08 & 31.32 & 62.21 \\
\cline{2-8}
\rule{0pt}{2ex} & \texttt{qwen2.5-0.5B-instr} + Pointwise & 52.47 & 46.12 & 41.67 & 33.09 & 42.89 & 65.51 \\
\rule{0pt}{2ex} & \texttt{qwen2.5-0.5B-instr} + Pointwise + Hints & 56.71 & 49.29 & 44.20 & 35.27 & 46.88 & 69.03 \\
& \texttt{qwen2.5-3B-instr} ~~~ + Pointwise & 58.15 & 48.56 & 42.01 & 31.44 & 44.42 & 68.82 \\
& \texttt{qwen2.5-3B-instr} ~~~ + Pointwise + Hints  & \textbf{64.39} & \textbf{55.07} & \textbf{47.99} & \textbf{35.98} & \textbf{50.24} & \textbf{74.74} \\
\bottomrule
\end{tabular*}
\caption{Ranking results of Claude Sonnet 3.5 v2 on the test set. The results show that a hint-augmented pointwise ranker achieves the best results across all metrics.
}
\label{tab:results}
\end{table*}

\paragraph{Dense Retriever}
For all experiments, we use the \texttt{Alibaba-NLP/gte-Qwen2-7B-instruct} embedding model as the first-stage retriever, to ensure fair comparison across re-ranking methods.

\subsection{Re-ranking Models}
\paragraph{Listwise}
Concatenating all 50 retrieved products (titles and descriptions) yields ~28k tokens, exceeding most LLM context windows and triggering the ``lost in the middle'' phenomenon \citep{liu-etal-2024-lost}. While pairwise rerankers often perform better \citep{qin-etal-2024-large}, they require $O(N)$ inference calls versus $O(1)$ for listwise approaches, creating substantial computational overhead. We address this trade-off through chunking: dividing 50 products into 5 chunks of 10, processing each chunk to extract the top 2 candidates with similarity scores, then merging results into the final ranking. This preserves listwise efficiency while mitigating context limitations. Given $\sim$300 tokens per product, each chunk requires $\sim$3.3k tokens; adding few-shot examples would reintroduce the ``lost in the middle'' problem we aim to avoid. Therefore, we employ zero-shot prompting.

We evaluate two state-of-the-art open-source LLMs: (1) a base model \texttt{Qwen2.5-72B-Instruct} \citep{qwen2.5}, and (2) a reasoning model \texttt{DeepSeek-R1} \citep{deepseek2025r1}. These models establish strong baselines for evaluating LLM performance on our reranking task.

\paragraph{Pointwise Finetuning}
While listwise in-context-learning (ICL) ranking effectively leverages LLMs' internal world knowledge, its computational demands present significant challenges. We thus train small language models (SLMs) with fewer than 3B parameters as a resource-efficient approach to incorporate world knowledge from larger models.
We implement a pointwise method that treats each query-product pair individually, using two model sizes from the \texttt{qwen2.5} family \citep{qwen2.5}: \texttt{qwen2.5-0.5B-instruct} and \texttt{qwen2.5-3B-instruct}. Our baseline models process superlative queries paired with products retrieved by a dense retriever without additional text.

We also explore whether SLMs learn linguistic features and world knowledge extracted from large models.
Our exploration includes two methods to incorporate interpretations of superlative terms: (1) replacing original queries with LLM-generated query reformulations; and (2) augmenting input text with relevant semantic features extracted from larger models.\footnote{The comparative performance of various query generation methods is presented in Appendix \ref{appendix:query_generation}.} To enhance domain-specific performance in e-commerce applications, we also  incorporate the top three brands identified in LLM-generated hints at both training and inference time, which helps establish stronger relationships between queries and products.

\subsection{Results}
\paragraph{Automatic Evaluation} 
Table \ref{tab:results} reveals distinct performance patterns across retrieval and ranking methods. Among sparse retrievers, doc2query shows modest improvement over vanilla BM25 (15.41 vs 14.91 P@1), demonstrating that generic query expansion provides limited benefits for superlative queries. In contrast, our QE-BM25 achieves 25.93 P@1, significantly outperforming BM25 by explicitly decomposing superlative terms into domain-specific attributes. This substantial gap (+10.5 points over doc2query) demonstrates that superlative-aware interpretation provides advantages beyond standard QE methods. The dense retriever achieves 20.14 p@1, falling between generic QE and our attribute-based approach.

For reranking, the listwise reranker \texttt{qwen2.5-72B-instruct} shows limited effectiveness (35.31 P@1), while the reasoning-focused \texttt{deepseek-r1} achieves superior results (47.06 P@1). Fine-tuned SLMs demonstrate exceptional efficiency: without interpretative hints (P), the 3B model achieves 58.15 p@1, surpassing both listwise approaches, and the 0.5B model reaches 52.47 P@1, outperforming the 72B model. When enhanced with LLM-generated superlative interpretations (P+H), performance increases substantially: the 3B model improves to 64.39 P@1 (+6.2 points) and the 0.5B model to 56.71 p@1 (+4.2 points), demonstrating the effectiveness of structured decomposition.

To avoid any judge-specific biases, we repeat our evaluation using Amazon Nova Pro \citep{amazon2024nova} as an alternative LLM judge in Appendix \ref{appendix:nova_results}, and confirm consistent relative performance patterns.

\paragraph{Human Evaluation}
From our test set, we randomly sample 153 queries and ask ten expert annotators to judge which of the two pointwise \texttt{qwen2.5-3B-instruct} models produced better top-10 ranked products, choosing between \textit{system\_a} (with hints), \textit{system\_b} (baseline), or \textit{tie}. The evaluation was conducted in a blind setting, with annotators unaware of which system was which. Inter-annotator agreement measured by Cohen's kappa is 0.74, %
indicating substantial agreement. Out of 153 assessed queries, the hint-augmented system outperforms the baseline in 51 cases, while the baseline wins in 27 cases, with 75 ties. Excluding ties, this represents a win rate of 65.38\% (48/78) for our proposed approach. These human judgments strongly align with our automatic evaluation results, confirming that the relevance labels generated through hints effectively transfer knowledge from larger models to our ranking system. We also conduct a qualitative analysis to further explore the results (see Appendix \ref{appendix_subsec:qualitative}).

\subsection{Efficiency}
\label{sec:efficiency}

While the ranking quality improvements demonstrated in Table \ref{tab:results} are substantial, practical deployment requires careful consideration of computational costs and latency. We evaluate efficiency from two complementary perspectives: hardware-independent computational cost (PFlops, in Appendix \ref{appendix_subsec:efficiency}) and real-world end-to-end latency.

\begin{table}[t!]
\centering\small
\begin{tabular}{l|l|ccc}
\toprule
\textbf{Retriever} & \textbf{Reranker} & \textbf{Avg} & \textbf{p5} & \textbf{p95} \\
\midrule
BM25 & N/A & 0.003 & 0.002 & 0.005 \\
QE-BM25 & N/A & 4.291 & 3.540 & 5.084 \\
\midrule
\multirow{5}{*}{\texttt{gte-qwen2-7b}} & N/A & 3.688 & 3.648 & 3.729 \\
& 0.5B P & 4.215 & 4.157 & 4.258 \\
& 0.5B P+H & 4.364 & 4.294 & 4.394 \\
& 3B P & 6.912 & 6.719 & 7.038 \\
& 3B P+H & 7.385 & 7.015 & 7.643 \\
\bottomrule
\end{tabular}
\caption{End-to-end latency per query (seconds) on a single L40 GPU. P: pointwise, H: hints. p5 and p95 denote 5th and 95th percentile latencies. Due to GPU memory constraints, we could not evaluate listwise LLM rerankers in this setup.}
\label{tab:latency}
\end{table}

\paragraph{End-to-End Latency}

Table \ref{tab:latency} reports latency including Claude-generated hints demonstrating relative overhead; in production, hints could be generated by lightweight local models. Dense retrieval using FAISS ANN search establishes a baseline of 3.688s per query. Our architecture enables hint generation to execute concurrently with retrieval; consequently, incorporating hints introduces minimal overhead: 149ms (3.5\%) for the 0.5B model and 473ms (6.8\%) for the 3B model. This overhead reflects longer input sequences rather than hint generation latency. When synthesized with quality metrics from Table \ref{tab:results}, the efficiency-quality trade-off proves compelling: our 0.5B+H model achieves 56.71 P@1 in 4.364s, outperforming \texttt{qwen2.5-72B-instruct} by 21.4 points, while the 3B+H model attains 64.39 P@1 in 7.385s, surpassing \texttt{deepseek-r1} by 17.3 points. Our models require 0.122–0.224 PFlops versus 1.200–1.720 PFlops for listwise baselines, a 14× efficiency gain (Appendix \ref{appendix_subsec:efficiency}). These findings indicate that explicit superlative interpretation via hints enables production-scale deployment while maintaining superior ranking performance.

\section{Conclusion}
In this paper, we addressed the challenge of interpreting superlatives in e-commerce search queries by constructing a specialized dataset and leveraging LLMs to generate reasoning. Our experiments demonstrated that explicitly capturing world knowledge about superlatives significantly improves search quality across both retrieval and ranking tasks by 10.9 points in MAP and 5.9 points in MRR respectively.
Our work shows that both semantic knowledge and reasoning capabilities from LLMs can be effectively distilled into more efficient systems, making sophisticated superlative interpretation feasible for real-time search environments.

\clearpage

\section*{Limitations}
First, our experimental scope is limited in model diversity. For listwise ranking experiments, we evaluate only two open-source LLMs without including recent commercial models like OpenAI's o1 \citep{openai2024openaio1card} or Claude 3.7 Sonnet \citep{claude3.7sonnet}. Our pointwise experiments are restricted to Qwen family models. Evaluating more architecturally diverse models such as Phi-4 \citep{abdin2024phi4technicalreport} would better validate our approach's effectiveness across different model designs. Second, our method addresses only English (particularly US) queries, constraining the broader applicability of our findings. Superlative interpretations likely differ across languages with varying grammatical structures. This linguistic limitation restricts the generalizability of our results to multilingual settings where superlatives might function differently.
The hint-based ranking approach can also be expanded to non-superlative categories, and could be used to diversify search results for different query sub-types, such as high consideration queries \cite{chen-etal-2024-identifying}.
Future work can also consider how this approach can integrate into conversational shopping systems that can include functionality for conversational search, question suggestion, and more \cite{li-etal-2025-wizard, 10.1145/3687273.3687293,10.1145/3626772.3661371}.

\bibliography{custom,anthology}

\clearpage
\appendix

\section{Experiments}
\subsection{Experiments Setup}
We conduct all pointwise model training and evaluation using NVIDIA L40s GPUs from AWS g6e.48xlarge instances (cost \$30.13 per hour as of July 2025). For listwise LLM inference, we use NVIDIA A100 GPUs from AWS p4de.24xlarge instances (cost \$27.45 per hour as of July 2025). All pointwise models are trained with early stopping based on validation performance and learning rate of 1e-5. The batch size for 0.5B model is 32 and for 3B is 2 per GPU. Inference is performed with half-precision (FP16) to optimize throughput while maintaining ranking quality.

\begin{table*}[t]
\centering\small
\begin{tabular*}{\textwidth}{@{\extracolsep{\fill}}l|l|cccccc}
\toprule
\textbf{Retriever} & \textbf{Reranker} & \textbf{p@1} & \textbf{p@3} & \textbf{p@5} & \textbf{p@10} & \textbf{MAP} & \textbf{MRR} \\
\midrule
bm25 & \multirow{4}{*}{N/A} & 22.44 & 21.84 & 22.07 & 22.28 & 37.26 & 42.52 \\
doc2query & & 26.68 & 25.42 & 25.26 & 24.61 & 40.31 & 46.15 \\
qe-bm25 & & \underline{31.78} & \underline{32.20} & \underline{32.05} & \underline{31.44} & \underline{46.15} & 52.78 \\
\texttt{gte-qwen2-7b-instr} & & 37.23 & 38.07 & 37.09 & 35.81 & 49.98 & \underline{57.02} \\
\midrule
\multirow{6}{*}{\texttt{gte-qwen2-7b-instr}} & \texttt{qwen2.5-72B-instr}, Listwise & 49.71 & 49.31 & 48.98 & 45.65 & 19.59 & 67.78 \\
& \texttt{deepseek-r1}, Listwise & 60.02 & 56.59 & 54.09 & 48.26 & 22.53 & 74.96 \\
\cline{2-8}
\rule{0pt}{2ex} & \texttt{qwen2.5-0.5B-instr}, Pointwise & 56.38 & 52.87 & 50.97 & 45.79 & 46.73 & 70.99 \\
\rule{0pt}{2ex} & \texttt{qwen2.5-0.5B-instr}, Pointwise + Hints & 66.77 & 59.99 & 55.82 & 48.46 & 49.44 & 78.20 \\
& \texttt{qwen2.5-3B-instr}, Pointwise & 67.64 & 59.57 & 53.98 & 45.07 & 47.79 & 78.14 \\
& \texttt{qwen2.5-3B-instr}, Pointwise + Hints  & \textbf{72.50} & \textbf{63.72} & \textbf{58.02} & \textbf{48.52} & \textbf{50.26} & \textbf{81.68} \\
\bottomrule
\end{tabular*}
\caption{Ranking results using Amazon Nova Pro as judge on the test set.}
\label{tab:nova_results}
\end{table*}

\subsection{Additional Experiments Results} \label{appendix:nova_results}
To address potential circularity concerns from using Claude 3.5 Sonnet v2 as both hint generator and relevance judge, we evaluated all methods using Amazon Nova Pro as an alternative judge. Table \ref{tab:nova_results} presents the complete results across all retrieval and ranking methods. While Nova Pro exhibits more lenient classification tendencies, particularly for ``best match'' assessments, resulting in higher absolute scores across all methods, the relative performance patterns remain consistent with our main results in Table \ref{tab:results}.

The Nova Pro evaluation reveals several key observations. First, all methods achieve higher absolute scores compared to Claude 3.5 Sonnet v2 judgments, reflecting Nova Pro's more lenient assessment criteria. For instance, the 3B+H model achieves 72.50 p@1 with Nova Pro versus 64.39 with Claude. Second, the relative improvements from hint augmentation remain consistent across judges. The 3B model gains +4.9 points (67.64 to 72.50) with Nova Pro compared to +6.2 points (58.15 to 64.39) with Claude, while the 0.5B model gains +10.4 points (56.38 to 66.77) with Nova Pro compared to +4.2 points (52.47 to 56.71) with Claude. Third, the ranking of methods remains stable: hint augmented models consistently outperform their non-hint counterparts, and our proposed 3B+H model achieves the best performance under both evaluation frameworks. These consistent relative patterns, combined with our human evaluation (Section 3.3), confirm that our findings are not artifacts of judge specific biases and that hint based superlative interpretation provides genuine improvements in ranking quality.

\section{Dataset}

\subsection{Superlative Queries} \label{appendix:example_query}
Table \ref{tab:superlative_queries} presents example superlative queries from the dataset.

\begin{table}[t]
\centering
\small
\begin{tabular}{p{0.45\textwidth}}
\toprule
\textbf{Query} \\
\midrule
best metallic balloons \\
most comfortable running shoes \\
top noise cancelling earphones \\
most durable luggage locks for international travel \\
best stamp ink pads that don't dry out \\
best paint pens for rock painting \\
best car washing sponge \\
leading esports gaming headsets \\
best microfiber towels for car detailing \\
\bottomrule
\end{tabular}
\caption{Example superlative queries from the dataset.}
\label{tab:superlative_queries}
\end{table}

\subsection{Dataset Quality} \label{appendix:data_quality}
To validate our generated superlative queries, we randomly selected 1,000 examples for human evaluation. An annotator assessed whether each query appeared natural and representative of authentic user search behavior in e-commerce contexts. The evaluation results showed that 946 queries (94.6\%) were judged as valid and realistic, indicating the satisfactory quality of our dataset for studying superlative expressions in search contexts.

\section{Query Enhanced BM25}
\subsection{Algorithm} \label{appendix_subsec:algorithm}

Algorithm \ref{algorithm:qe_bm25} details our implementation of QE-BM25, which enhances retrieval by averaging BM25 scores across multiple query reformulations. In particular, for each original query, we calculate BM25 scores for all its variations against the document corpus, then rank documents by their average relevance across variations. This approach helps address vocabulary mismatch issues, particularly for superlative and comparative queries.

\begin{algorithm}[h!]
\caption{Query Enhanced BM25 (QE‑BM25)}
\label{algorithm:qe_bm25}
\scriptsize
\begin{algorithmic}[1]
  \Require $Q = \{q_1,\dots,q_n\}$  \Comment{original user queries}
  \Require $QG = \{\mathit{QG}_1,\dots,\mathit{QG}_n\}$ with $\mathit{QG}_i=\{qg_{i1},qg_{i2},\dots\}$  \Comment{generated variants}
  \Require Corpus $\mathcal{D}$ of size $C$; top‑$k$ documents to return; $\texttt{max\_candidates}$
  \Ensure  $\texttt{finalScores},\texttt{finalDocIds}\in\mathbb{R}^{k\times n}$

  \State $\texttt{finalScores}\gets \mathbf{0}_{k\times n}$;\;
        $\texttt{finalDocIds}\gets \mathbf{0}_{k\times n}$
  \For{$i\gets 1$ to $n$}
      \State $\texttt{queryVars}\gets QG_i$
      \State $C_{\text{eff}}\gets\min\bigl(C, \texttt{max\_candidates}\bigr)$
      \State $\texttt{tempScores}\gets \mathbf{0}_{C_{\text{eff}}\times |\texttt{queryVars}|}$
      \For{$j\gets 1$ to $|\texttt{queryVars}|$}
          \State $\mathit{gq}\gets \texttt{queryVars}[j]$
          \State $\texttt{bm25}\gets\textsc{CalculateBM25}(\mathit{gq}, \mathcal{D}, \texttt{max\_candidates})$
          \ForAll{$(d, s)\in\texttt{bm25}$}
              \State $\texttt{tempScores}[d,j]\gets s$
          \EndFor
      \EndFor
      \State $\texttt{cands}\gets [\,]$
      \For{$d\gets 1$ to $C_{\text{eff}}$}
          \If{$\textsc{HasNonZeroValue}\!\bigl(\texttt{tempScores}[d,:]\bigr)$}
              \State $avg\gets\operatorname{mean}\bigl(\texttt{tempScores}[d,:]\bigr)$
              \State $\texttt{cands} \leftarrow \texttt{cands} \cup \{(d,avg)\}$
          \EndIf
      \EndFor
      \State $\mathit{top}\gets \textsc{SortDescending}(\texttt{cands})_{1..k}$
      \For{$r\gets 1$ to $\min(k,|\mathit{top}|)$}
          \State $(d,s)\gets\mathit{top}[r]$
          \State $\texttt{finalScores}[r,i]\gets s$
          \State $\texttt{finalDocIds}[r,i]\gets d$
      \EndFor
  \EndFor
  \State \Return $\bigl(\texttt{finalScores},\,\texttt{finalDocIds}\bigr)$
\end{algorithmic}
\end{algorithm}

\subsection{Coverage Analysis} \label{appendix_subsec:coverage}
Table \ref{tab:bm25_anlaysis} shows the coverage efficiency of our QE-BM25 implementation at different candidate thresholds. Average coverage measures the percentage of relevant documents captured within the top-k candidates, while perfect coverage indicates the percentage of queries for which all top-10 relevant documents are found. The analysis reveals that average coverage reaches 100\% at 50k candidates, while perfect coverage plateaus at 95.60\% from 40k candidates onward. This suggests that approximately 4.4\% of queries have at least one relevant document that falls outside our candidate pool, even at higher thresholds. Based on these results, we selected 50,000 as our \texttt{max\_candidates} parameter to maximize average coverage while maintaining reasonable computational efficiency.

\begin{table}[ht]
    \begin{tabular}{r|r|r}
    \toprule
    Search k & Avg. Coverage & Perfect Coverage \\
    \hline
    5k     & 99.35\%      & 94.00\%           \\
    10k    & 99.45\%      & 94.50\%           \\
    20k    & 99.58\%      & 95.90\%           \\
    30k    & 99.54\%      & 95.50\%           \\
    40k    & 99.56\%     & 95.60\%          \\ 
    50k    & 100.00\%     & 100.00\%          \\ 
    \bottomrule
    \end{tabular}
    \caption{BM25 coverage analysis showing the percentage of relevant documents captured (Avg. Coverage) and percentage of queries with all top-10 relevant documents found (Perfect Coverage) at various candidate pool sizes. Analysis performed on 1,000 sample queries.}
    \label{tab:bm25_anlaysis}
\end{table}

\section{Analysis} \label{appendix:analysis}
\subsection{Qualitative Evaluation} \label{appendix_subsec:qualitative}
To better understand how our hint-based approach improves model performance, we conducted a detailed qualitative analysis on 150 randomly selected queries. Through careful examination of system outputs, we identified three major error categories that frequently occurred in the baseline model but were substantially reduced in our hint-augmented approach. As presented in Table \ref{tab:analysis}, these error categories comprise: (1) brand recognition errors, where the model fails to properly interpret brand preferences implied by superlatives - as seen in the \textit{luxury sandals} example where the baseline recommends OluKai (a mid-tier brand) while the hint-augmented model correctly identifies Tory Burch as a luxury option; (2) feature interpretation errors, demonstrated in the documentary camera example where the baseline suggests a consumer-grade Fujifilm camera while our approach correctly prioritizes Sony's professional camcorder with features suitable for documentary work; and (3) relevancy assessment errors, illustrated by the tablecloth query where the baseline model recommends a general \textit{6 ft} tablecloth rather than one matching the precise \textit{60$\sim$102} dimensions specified in the query. The generated hints effectively address these error types by making implicit knowledge explicit.

\subsection{Efficiency} \label{appendix_subsec:efficiency}
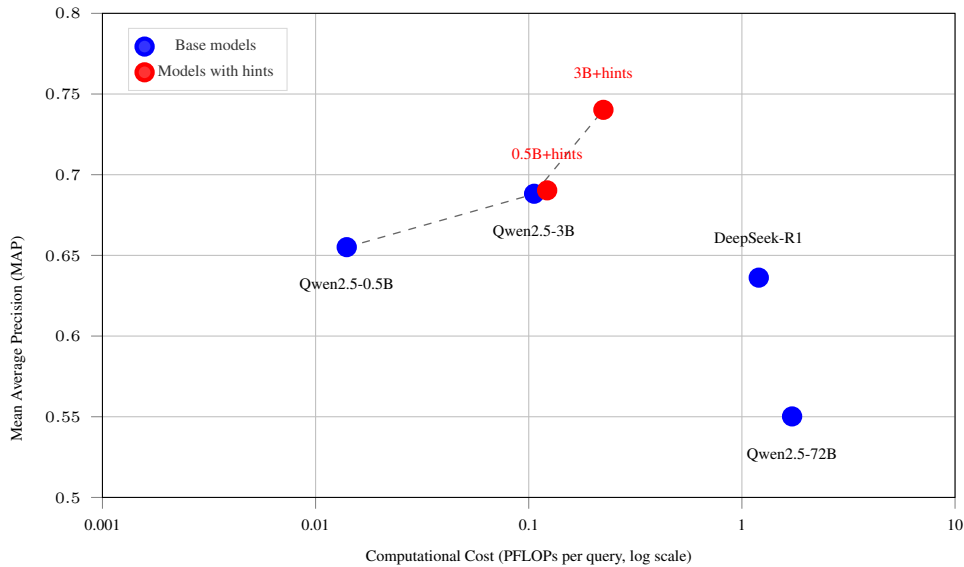
\begin{figure*}
\centering
\begin{tikzpicture}
\tikzset{every node/.style={font=\tiny,align=center}}

\begin{axis}[
    xlabel={Computational Cost (PFLOPs per query, log scale)},
    ylabel={Mean Average Precision (MAP)},
    xmode=log, log basis x={10},
    xmin=0.001, xmax=10,
    xtick={0.001,0.01,0.1,1,10},
    xticklabels={0.001,0.01,0.1,1,10},
    minor x tick num=8,
    ymin=0.50, ymax=0.80,
    grid=both,
    legend pos=north west,
    width=0.8\textwidth,
    height=8cm,
    tick align=outside,
    tick pos=left,
    every axis plot/.append style={ultra thick},
    legend style={font=\small, draw=gray!30, fill=white, fill opacity=0.8},
    ylabel style={xshift=-1.1cm},
]

\addplot[blue, mark=*, mark size=3pt, only marks]
coordinates {
    (0.014, 0.6551)   %
    (0.106, 0.6882)   %
    (1.720, 0.5502)   %
    (1.200, 0.6362)   %
};

\addplot[red, mark=*, mark size=3pt, only marks]
coordinates {
    (0.122, 0.6903)   %
    (0.224, 0.7402)   %
};

\addplot[dashed,->,color=black!60,line width=0.5pt]
coordinates {(0.014,0.6551) (0.122,0.6903)};

\addplot[dashed,->,color=black!60,line width=0.5pt]
coordinates {(0.106,0.6882) (0.224,0.7402)};

\legend{Base models, Models with hints};

\node[anchor=north ,yshift=-8pt] at (axis cs:0.014,0.6551) {Qwen2.5-0.5B};
\node[anchor=north ,yshift=-8pt] at (axis cs:0.106,0.6882) {Qwen2.5-3B};
\node[anchor=north ,yshift=-8pt] at (axis cs:1.720,0.5502) {Qwen2.5-72B};
\node[anchor=south ,yshift= 8pt] at (axis cs:1.200,0.6362) {DeepSeek-R1};
\node[text=red,anchor=south,yshift=8pt] at (axis cs:0.122,0.6903) {0.5B+hints};
\node[text=red,anchor=south,yshift=8pt] at (axis cs:0.224,0.7402) {3B+hints};

\end{axis}
\end{tikzpicture}
\caption{Relationship between computational cost and MRR performance.}
\label{fig:flops_vs_map}
\end{figure*}

Figure~\ref{fig:flops_vs_map} presents how adding lightweight retrieval hints (red) consistently boosts MRR scores at marginal additional computational cost, allowing sub‑billion‑parameter Qwen2.5 models to outperform the strong baselines while remaining two–three orders of magnitude cheaper in PFLOPs per query.

Our approach achieves computational efficiency through key framework design. First, the fine-tuning cost of our lightweight models is minimal, requiring only 3–4 hours on a single NVIDIA L40s GPU—a one-time expense amortized across many inference operations. This upfront training enables substantial long-term computational savings compared to repeatedly invoking large models for every query at inference time.
Second, the inference overhead associated with hint generation can be significantly reduced by precomputing and caching hints for frequently encountered superlative queries. This offline preprocessing effectively minimizes latency for common query patterns, addressing practical concerns around online latency.
Finally, our pipeline design supports concurrent execution of hint generation and retrieval, greatly diminishing the actual latency impact. By allowing these two steps to run in parallel, our method maintains competitive ranking effectiveness while satisfying stringent real-world latency constraints.

\subsection{Query Generation Methods} \label{appendix:query_generation}
We evaluate two primary query generation approaches: (1) replacing original queries with LLM-generated structured attribute-value list and (2) augmenting input text with semantic features extracted from larger models. Then, we append brand information as additional context lines. This enriched text is formatted as \textit{relevance query: \{enriched\_query\} \{brand\_info\} product: \{product\_details\}} before tokenization and input to the pointwise reranker. Table \ref{tab:query_generation} presents comparative results using \texttt{qwen2.5-3B-instruct}, showing that while feature-based augmentation performs marginally better at p@1, query reformulations demonstrate consistent improvements across p@3, p@5, p@10, MAP, and MRR metrics, indicating better overall ranking performance.

\begin{table*}[h]
    \centering\small
    \begin{tabularx}{\textwidth}{p{0.03\textwidth}|p{0.45\textwidth}|p{0.45\textwidth}}
    \toprule
    Rank & Baseline\_3B & 3B+hints \\
    \midrule
    1 & OluKai Kaekae   Women's Beach Sandals, Full-Grain \& Metallic Leather Flip-Flop Slides   with Wet Grip Soles, Lightweight \& All-Day Comfort & Tory Burch Women's   Emmy Flat Sandals, Ambra, Tan, 7 Medium US                    \\
    \hdashline
    \rule{0pt}{2ex}2 & OluKai Ho'opio Leather Women's Beach   Sandals, Full-Grain Leather Flip-Flop Slides, Modern Low Profile Design \&   Comfortable Fit          & Tory Burch Women's Double T Sport Slides,   Spark Gold/New Cream/Gold, 6 Medium US \\
    \hdashline
    \rule{0pt}{2ex}3 & Tory Burch Women's Emmy Flat Sandals,   Ambra, Tan, 7 Medium US                                                                              & M by Bruno Magli Womens Jade Sandal (9 B   US, Beige)                              \\
    \hdashline
    \rule{0pt}{2ex}4 & Tory Burch Women's Double T Sport Slides,   Spark Gold/New Cream/Gold, 6 Medium US                                                           & Tory Burch Women's Miller Thong Sandals,   Black, 7.5 Medium US                    \\
    \hdashline
    \rule{0pt}{2ex}5 & Tory Burch Women's INES Slide Leather   Cushioned Sandals Black Size 5.5                                                                     & Tory Burch Women's Miller Thong Sandals,   Black, 8.5 Medium US                    \\
    \hdashline
    \rule{0pt}{2ex}6 & ory Burch Women's Miller Thong Sandals,   Black, 10.5 Medium US                                                                             & Tory Burch Women's Miller Thong Sandals,   Black, 7 Medium US                      \\
    \hdashline
    \rule{0pt}{2ex}7 & Tory Burch Women's Miller Thong Sandals,   Black, 7 Medium US                                                                                & Tory Burch Women's Miller Thong Sandals,   Black, 10.5 Medium US                   \\
    \hdashline
    \rule{0pt}{2ex}8 & Tory Burch Women's Miller Thong Sandals,   Black, 7.5 Medium US                                                                              & Tory Burch Women's INES Slide Leather   Cushioned Sandals Black Size 5.5           \\
    \hdashline
    \rule{0pt}{2ex}9 & Tory Burch Women's Miller Thong Sandals,   Black, 8.5 Medium US                                                                              & Tory Burch Women's Manon Thong Sandals,   Perfect Black, 6.5 Medium US             \\
    \hdashline
    \rule{0pt}{2ex}10 & Tory Burch Women's Manon Thong Sandals,   Perfect Black, 6.5 Medium US                                                                       & Chaco Women's Zcloud X2 Sandal, Hermes   Gold, 8 \\
    \bottomrule
\end{tabularx}
    \caption{Example output of the reranked lists for query \textit{best women's luxury sandals}.}
    \label{tab:example}
\end{table*}

Table \ref{tab:example} presents comparative ranking examples from both our baseline model and hint-augmented model. These examples illustrate how our approach effectively interprets superlative expressions to produce more relevant product rankings aligned with implicit user preferences.

\subsection{Category-based Analysis}
\label{appendix:category}

Table~\ref{tab:category_performance} presents the ranking performance evaluated by Claude Sonnet 3.5 v2 across 25 product categories. We focus our analysis on categories with sufficient query volume (queries $\geq$ 10) to ensure statistical reliability, which accounts for 99.81\% of the total evaluation set. The model demonstrates relatively stable performance across categories, with MRR scores ranging from 64.07 to 81.73 for major categories. 

\begin{table*}[thb]
\centering
\small
\begin{tabular*}{\textwidth}{@{\extracolsep{\fill}}lrrrrrrrrr@{}}
\toprule
\textbf{Category} & \textbf{\%} & \textbf{Queries} & \textbf{P@1} & \textbf{P@3} & \textbf{P@5} & \textbf{P@10} & \textbf{MAP} & \textbf{MRR} \\
\midrule
Clothing, Shoes \& Accessories & 25.87 & 1384 & 61.05 & 51.61 & 44.55 & 32.70 & 48.05 & 71.10 \\
Home \& Garden & 19.94 & 1067 & 62.70 & 55.23 & 48.51 & 36.81 & 48.04 & 74.56 \\
Health \& Personal Care & 12.54 & 671 & 66.77 & 56.78 & 48.79 & 36.53 & 49.44 & 76.83 \\
Electronics \& Computers & 11.59 & 620 & 73.87 & 64.62 & 57.65 & 43.21 & 57.99 & 81.73 \\
Sports \& Outdoors & 5.59 & 299 & 63.88 & 55.63 & 48.23 & 36.62 & 53.08 & 75.32 \\
Toys \& Games & 5.27 & 282 & 70.92 & 58.04 & 50.43 & 38.37 & 55.84 & 80.60 \\
Automotive \& Marine & 3.50 & 187 & 59.89 & 47.42 & 40.64 & 29.63 & 50.55 & 69.69 \\
Jewelry \& Watches & 3.25 & 174 & 59.20 & 50.19 & 43.33 & 32.47 & 47.15 & 69.70 \\
Food \& Beverages & 2.84 & 152 & 65.79 & 56.36 & 49.87 & 39.47 & 53.92 & 76.97 \\
Office \& School Supplies & 2.34 & 125 & 59.20 & 53.87 & 46.88 & 37.76 & 46.84 & 71.30 \\
Pet Supplies & 1.61 & 86 & 61.63 & 57.75 & 53.72 & 41.74 & 53.74 & 74.11 \\
Tools \& Home Improvement & 1.51 & 81 & 66.67 & 52.26 & 45.68 & 32.72 & 50.77 & 76.64 \\
Furniture \& Decor & 1.18 & 63 & 61.90 & 47.09 & 38.73 & 25.40 & 40.25 & 72.00 \\
Arts, Crafts \& Party Supplies & 1.16 & 62 & 51.61 & 41.40 & 34.84 & 25.97 & 40.53 & 64.07 \\
Bags \& Luggage & 0.90 & 48 & 70.83 & 59.72 & 51.25 & 38.54 & 51.80 & 79.43 \\
Baby \& Child Care & 0.58 & 31 & 70.97 & 52.69 & 47.74 & 36.13 & 48.09 & 79.49 \\
\bottomrule
\end{tabular*}
\caption{Performance Metrics by Category of model \texttt{qwen2.5-3B-instruct P+H}. Categories with less than 10 queries are dropped.}
\label{tab:category_performance}
\end{table*}

\textsc{Electronics \& Computers} achieves the highest MRR at 81.73, followed by \textsc{Toys \& Games} at 80.60 and \textsc{Baby \& Child Care} at 79.49, while \textsc{Automotive \& Marine} at 69.69, \textsc{Jewelry \& Watches} at 69.70, and \textsc{Arts, Crafts \& Party Supplies} at 64.07 show relatively lower performance. Notably, the model exhibits consistent performance across categories of varying popularity: mid-tail categories such as \textsc{Sports \& Outdoors} with 5.59\% of queries achieving MRR of 75.32, and \textsc{Food \& Beverages} with 2.84\% achieving MRR of 76.97, perform comparably to the most populated categories like \textsc{Clothing, Shoes \& Accessories} with 25.87\% achieving MRR of 71.10, and \textsc{Home \& Garden} with 19.94\% achieving MRR of 74.56. This consistency suggests that the model generalizes well across different product domains regardless of category prevalence, which is crucial for real-world e-commerce applications.

\subsection{Long-tail Query Analysis}
\label{appendix:longtail}

To evaluate robustness across varying query difficulty levels, we analyze performance on queries with different numbers of relevant items. Table \ref{tab:longtail} presents results stratified by the number of relevant items in the test set, using our best performing model (\texttt{qwen2.5-3B-instruct} P+H).

The stratified analysis reveals expected performance variations based on query difficulty. Queries with fewer relevant items exhibit lower performance metrics, with p@1 ranging from 35.97 to 40.17 compared to 64.39 overall, reflecting the increased challenge when relevant candidates are sparse. Nevertheless, the model maintains substantial effectiveness across all subsets, with MRR values between 46.42 and 52.52 (versus 74.74 overall) indicating that relevant items consistently appear in top ranking positions. Performance scales monotonically with the number of relevant items: p@1 increases from 35.97 (one item) to 40.17 (three items), and MRR from 46.42 to 52.52. MAP values remain stable across subsets (43.04 to 46.42 versus 50.24 overall), demonstrating moderate rather than severe degradation on challenging queries. These results confirm that our approach exhibits robust performance across varying query difficulty levels, maintaining practical utility even for long tail queries with sparse relevant items.

\begin{table*}[h]
    \small
    \setlength{\tabcolsep}{4pt}
    \begin{tabularx}{\textwidth}{p{0.15\textwidth}|p{0.35\textwidth}|p{0.35\textwidth}|p{0.15\textwidth}}
    \toprule
    \textbf{Query} & \textbf{Baseline} & \textbf{Hint-augmented} & \textbf{Category} \\
    \midrule
    best women's luxury sandals & OluKai Kaekae Women's Beach Sandals, Full-Grain \& Metallic Leather Flip-Flop Slides with Wet Grip Soles, Lightweight \& All-Day Comfort & Tory Burch Women's Emmy Flat Sandals, Ambra, Tan & Brands \\
    \hline
    \rule{0pt}{2ex}best rectangle table cloths 60$\times$102 & Surmente Table Cloth 6 ft Rectangular Polyester Tablecloth Tablecloths for Rectangle Tables for Weddings, Banquets, or Restaurants (Red) & Kadut Rectangle Tablecloth (60 x 102 Inch) Beige Rectangular Table Cloth for 6 Foot Table | Heavy Duty Fabric | Stain Proof Table Cloth for Parties, Weddings, Kitchen, Wrinkle-Resistant Table Cover & Relevancy \\
    \hline
    \rule{0pt}{2ex}most popular digital cameras for documentary & Fujifilm X-T30 Mirrorless Digital Camera, Charcoal Silver with Fujinon XC15-45mm Optical Image Stabilisation Power Zoom Lens kit, Black & Sony PXW-FS5 XDCAM Super 35 Camera System Professional Camcorder, Black (PXWFS5) & features \\
    \bottomrule
    \end{tabularx}
    \caption{Examples showing how hint augmentation improves product ranking quality across different types of superlative search queries.}
    \label{tab:analysis}
\end{table*}

\begin{table}[h]
    \centering\small
    \begin{tabular}{l|llllll}
    \toprule
    Type & p@1   & p@3   & p@5   & p@10  & MAP   & MRR   \\
    \midrule
    Features & 64.47 & 53.75 & 46.90 & 35.17 & 49.17 & 74.02 \\
    Queries & 64.39 & 55.07 & 47.99 & 35.98 & 50.24 & 74.74 \\
    \bottomrule
    \end{tabular}
    \caption{Comparison between two query generation methods.}
    \label{tab:query_generation}
\end{table}

\begin{table*}[t]
\centering\small
\begin{tabular*}{\textwidth}{@{\extracolsep{\fill}}l|cc|ccccccc}
\toprule
\textbf{\# Relevant} & \textbf{\# Queries} & \textbf{\# Pairs} & \textbf{p@1} & \textbf{p@3} & \textbf{p@5} & \textbf{p@10} & \textbf{MAP} & \textbf{MRR} \\
\midrule
1 & 467 & 23,350 & 35.97 & 17.20 & 11.52 & 6.55 & 46.42 & 46.42 \\
2 & 883 & 44,150 & 37.26 & 21.37 & 15.15 & 8.97 & 43.74 & 48.73 \\
3 & 1,277 & 63,850 & 40.17 & 25.45 & 18.68 & 11.41 & 43.04 & 52.52 \\
\midrule
Overall & 5,353 & 267,650 & 64.39 & 55.07 & 47.99 & 35.98 & 50.24 & 74.74 \\
\bottomrule
\end{tabular*}
\caption{Performance breakdown by number of relevant items per query for \texttt{qwen2.5-3B-instruct} P+H model. Results show consistent performance improvements as more relevant items become available, though queries with only one relevant item remain challenging.}
\label{tab:longtail}
\end{table*}

\section{Prompts}
\label{appendix:prompts}

We carefully designed prompts for dataset construction and interpretation of superlative words. Table \ref{tab:qeury_generation_prompt} shows the prompt for generating diverse superlative queries across product categories for dataset construction. Table \ref{tab:label_generation_prompt} presents the prompt used to create gold standard relevance labels for query-product pairs. Table \ref{tab:hint_generation_prompt} contains the prompt for generating hints that explain the interpretation of superlative terms in queries, which is a key component of our proposed method for handling superlative search queries. Table \ref{tab:listwise_prompt} presents the prompt used for listwise reranking.

\begin{table*}[ht]
    \centering\footnotesize
    \aboverulesep=0ex
    \belowrulesep=0ex
    
    \begin{tabularx}{\textwidth}{|X|}
    \hline
    \rule{0pt}{4ex}QUERY\_GENERATION\_PROMPT = """You are an e-commerce search expert. Generate \{n\} distinct, natural search queries for \{noun\} as found in real online shopping behavior.\\\\

\#\# Rules:\\
- Use superlative words: best, top, most popular, leading, etc.\\
- Make queries sound natural like actual customer searches\\
- Use the reference examples to generate queries with similar styles\\\\

\#\# Reference examples for various categories\\
- most popular electric shavers\\
- best black coffee\\
- best hair dryer brush set\\
- best smart tv\\
- best mens blue jeans\\
- best toenail cutters\\
- top running shoes for men\\
- best action movies\\
- best switch games for kids\\
- leading teether for baby\\
- best pc strategy games\\
- top laundry sanitizer for babies\\
- best cartridge for hp printer\\
- most popular mens smart dress watches\\
- famous gps for hiking\\
- best usb speakers\\
- best stylish travel walking shoes for women\\
- top air purifier with reusable filter\\
- best selling romance novels\\
- best jacket for winter\\
- famous brand sunglasses\\
- best natural shampoo without chemicals\\\\

\#\# Output\\
Return a Python list of {n} generated queries as strings, ordered by estimated search popularity (most popular first).\\\\

Your output: """
    \\\\
    \hline
    \end{tabularx}
    \caption{Prompt used for superlative query generation.}
    \label{tab:qeury_generation_prompt}
\end{table*}

\begin{table*}[ht]
    \centering\scriptsize
    \aboverulesep=0ex
    \belowrulesep=0ex
    
    \begin{tabularx}{\textwidth}{|X|}
    \hline
    \rule{0pt}{4ex}RELEVANCE\_ANNOTATION\_PROMPT = """You are an expert e-commerce search evaluator. Your task is to assess the relevance between search queries and products, with special attention to superlative qualifiers like "best" or "top" in the queries.\\\\

\#\# Understanding Superlative Product Queries\\
Superlative qualifiers in queries (like "best," "top," "most popular") indicate the user is seeking products that excel in specific dimensions:\\
- Quality superlatives ("best," "top," "finest"): Products with superior quality, performance, and features; typically premium or flagship offerings from respected brands\\
- Popularity superlatives ("most popular," "best-selling"): Products with widespread adoption, high sales volume, or strong social proof\\
- Rating superlatives ("highest rated," "best-reviewed"): Products with exceptional user or expert reviews\\
- Recognition superlatives ("leading," "famous"): Products or brands with established reputation and recognition\\
- Value superlatives ("best value," "top bang for buck"): Products offering optimal balance between price and quality/features\\
- Specialization superlatives ("best for X purpose," "top in Y environment"): Products that excel in specific use cases or contexts\\\\

\#\# Evaluation Steps\\
1. Identify if the query contains superlative qualifiers (e.g., "best", "top", "greatest").\\
2. Determine if the product meets the query's basic category and functional requirements.\\
3. If superlatives are present, assess if the product is genuinely recognized as among the top-tier options in its category based on:\\
   - Brand reputation and market position\\
   - Sales data and market share (for popularity claims)\\
   - Professional reviews and expert opinions\\
   - User ratings and community sentiment\\
   - Feature set and specifications relative to competitors\\
   - Quality of materials and construction\\
   - Price-to-performance ratio and value proposition\\
   - Special features that address niche requirements\\
   - Historical significance or innovation in the category\\
   - Awards, certifications, or industry recognition\\\\

\#\# Relevance Categories\\
- "relevant and best": Product meets basic requirements AND genuinely excels in at least one dimension implied by the superlative qualifier (not necessarily all dimensions). This could be through superior features, exceptional brand reputation, market leadership, outstanding user ratings, or specialized excellence in the specific context implied by the query.\\
- "relevant but not best": Product meets basic requirements but doesn't stand out in any dimension implied by the superlative qualifier compared to competing products in the same category.\\
- "irrelevant": Product doesn't meet basic category or functional requirements of the query.\\\\

\#\# Confidence Scoring Guidelines\\
When assigning a confidence score (0-100), consider:\\
- 80-100: High confidence - Clear evidence supports your assessment with minimal ambiguity\\
- 50-80: Moderate confidence - Some evidence exists but with limitations or potential counterarguments\\
- 0-50: Low confidence - Limited information, conflicting evidence, or high uncertainty\\\\

Factors affecting confidence include:\\
- Completeness of product information\\
- Clarity of the query's intent\\
- Availability of comparative data for superlative assessment\\
- Consistency of evidence across different evaluation factors\\\\

\#\# Input Format\\
You will receive a list of query-product pairs. Each pair consists of a search query and a product description. The input will contain **<batch\_size>** query-product pairs.\\\\

\#\# Output Format\\
You MUST provide exactly **<batch\_size>** evaluations in your output - one for each query-product pair in the input, maintaining their original order. Do not skip any pairs or combine evaluations.\\\\

Provide a list of dictionaries, one for each query-product pair, with each dictionary having the following format:\\
\{\\
  "reasoning": "Your detailed analysis of whether the product meets basic requirements and how it performs against superlative expectations. Include specific product attributes, market positioning, and comparative assessment that support your conclusion.",\\
  "label": "relevant and best" or "relevant but not best" or "irrelevant",\\
  "confidence": [number between 0-100]\\
\}

\#\# Input\\
<input>\\\\

\#\# Output"""
    \\\\
    \hline
    \end{tabularx}
    \caption{Prompt used for relevance label annotation.}
    \label{tab:label_generation_prompt}
\end{table*}

\begin{table*}[ht]
    \centering\scriptsize
    \aboverulesep=0ex
    \belowrulesep=0ex
    
    \begin{tabularx}{\textwidth}{|X|}
    \hline
    \rule{0pt}{4ex}HINT\_GENERATION\_PROMPT = """\# Brand and Feature Extraction from Superlative Queries with Synonyms\\
You are tasked with extracting relevant brands and features from user queries containing superlative terms (like "best," "top," "greatest," "most popular," etc.) to improve search result ranking for recommendation-seeking queries.\\
\\
\#\# Process\\
When presented with a superlative query, please:\\
1. Identify the product/service domain and specific category\\
2. Analyze what the superlative terms imply about user preferences (e.g., "best" might imply quality, "most popular" implies social proof)\\
3. Extract 5-10 most relevant brands, with confidence scores for each\\
4. Extract 5-10 most relevant features that would determine excellence in this category\\
5. Identify which brands are particularly known for excelling in these key features\\
6. For each feature, generate a list of synonyms and alternative phrasings to enable lexical matching\\
\\
\#\# Output Format\\
<analysis>\\
\{\\
    "domain": "Product/service category",\\
    "ranking\_intent": "Analysis of what the superlative qualifiers suggest about ranking criteria",\\
    "query\_clarification": "Any ambiguities that might benefit from clarification"\\
\}\\
</analysis>\\
\\
<brands>\\
\lbrack\\
    \{"name": "Brand 1", "confidence": 95\},\\
    \{"name": "Brand 2", "confidence": 90\},\\
    \# Include 5-10 brands total\\
\rbrack\\
</brands>\\
\\
<features>\\
\lbrack\\
    \{\\
        "name": "Feature 1", \\
        "synonyms": ["alternative term 1", "alternative term 2", "similar phrase", "related concept"],\\
        "category": "physical|performance|convenience|aesthetic|...", \\
        "importance": 10, \\
        "brands\_known\_for": ["Brand X", "Brand Y"]\\
    \},\\
    \{\\
        "name": "Feature 2", \\
        "synonyms": ["alternative term 1", "alternative term 2", "similar phrase"],\\
        "category": "physical|performance|convenience|aesthetic|...", \\
        "importance": 9, \\
        "brands\_known\_for": ["Brand Z"]\\
    \},\\
    \# Include 5-10 features total, importance scale 1-10\\
\rbrack\\
</features>\\
\\
<feature\_coverage\_queries>\\
\lbrack\\
    "Query incorporating ALL features using synonym set 1 for each feature", \\
    "Query incorporating ALL features using synonym set 2 for each feature", \\
    \# Include 10 queries total\\
\rbrack\\
</feature\_coverage\_queries>\\
\\
\#\# Guidelines\\
- Focus specifically on superlative queries seeking "best," "top," "highest rated," "most popular" items\\
- Provide 5-10 brands and 5-10 features (more for common queries, fewer for niche ones)\\
- Assign confidence scores (0-100) to brands based on their relevance to the superlative query\\
- Categorize features and assign importance weights (1-10) based on how critical they are to achieving excellence in this category\\
- Generate exactly <num\_queries> feature coverage queries where:\\
    - EACH query includes ALL identified features\\
    - Each query uses DIFFERENT synonyms for expressing the same features\\
    - Each query is phrased as a complete search query a user might type\\
    - Each query includes the core product/category concept from the original query\\
- Include common variations, industry-specific terminology, and consumer language in the synonyms\\
- For highly specialized queries, include closely related brands/alternatives\\
- Identify which brands are particularly known for excellence in which key features\\
- Format must be valid Python literal that can be parsed by ast.literal\_eval\\
\\
\#\# Query\\
<query>\\
\\
Your output: """
    \\\\
    \hline
    \end{tabularx}
    \caption{Prompt used for hint generation.}
    \label{tab:hint_generation_prompt}
\end{table*}

\begin{table*}[ht]
    \centering\scriptsize
    \aboverulesep=0ex
    \belowrulesep=0ex
    
    \begin{tabularx}{\textwidth}{|X|}
    \hline
    \rule{0pt}{4ex}BASELINE\_CHUNK\_RANKING\_PROMPT = """You are an expert e-commerce search evaluator. Your task is to identify the most relevant products from this subset for a given query, with particular attention to superlative qualifiers.\\
\\
Query: <query>\\
\\
Consider the following aspects when evaluating:\\
1. Relevance to the query's basic product/category requirement\\
2. Alignment with the superlative qualifier (e.g., "best", "top", "most popular")\\
3. Product features and how they relate to potential user intent\\
4. Brand reputation and popularity in the specific product category\\
5. Product reviews and ratings (if available)\\
6. Price point in relation to the query (if applicable)\\
\\
IMPORTANT INSTRUCTIONS:\\
- Return the **2 most relevant products** from the list below\\
- For each product, provide a similarity score from 0-100 (where 100 means perfect match to query)\\
- **Try to avoid numbers ending in 0/5 AND repetitive scores.**\\
- Your scores should reflect ABSOLUTE relevance to the query, not just relative ranking within this subset\\
- This ensures scores can be compared across different product subsets\\
- You MUST ONLY include product IDs that appear in this subset's list\\
- Your response MUST be a valid Python dictionary containing product IDs and their similarity scores\\
- Answer concisely: **Limit reasoning/thinking step (before </think>) to 5 sentences or fewer**\\
\\
Products:\\
<products>\\
\\
OUTPUT FORMAT:\\
Return ONLY a Python dictionary with exactly 2 product IDs and their similarity scores (0-100), in descending order of relevance:\\
\{\\
    "product\_id\_1": int,\\
    "product\_id\_2": int\\
\}\\
"""
    \\\\
    \hline
    \end{tabularx}
    \caption{Prompt used for list-wise LLM ranker.}
    \label{tab:listwise_prompt}
\end{table*}

\end{document}